\documentclass{article}
\usepackage{ijcai24}

\usepackage{times}
\usepackage{soul}
\usepackage{url}
\usepackage{xcolor}
\usepackage[hidelinks]{hyperref}
\usepackage[utf8]{inputenc}
\usepackage[small]{caption}
\usepackage{graphicx}
\usepackage{bbm}
\usepackage{amsmath,amsfonts}
\usepackage{amsthm}
\usepackage{booktabs}
\usepackage{algorithm}
\usepackage{algpseudocode}
\usepackage{svg}
\usepackage{subcaption}
\usepackage[switch]{lineno}

\newcommand{\cmmnt}[1]{}
\newcommand{\red}{\textcolor{red}}

\DeclareMathOperator*{\argmax}{arg\,max}

\algnewcommand\algorithmicforeach{\textbf{for each}}
\algdef{S}[FOR]{ForEach}[1]{\algorithmicforeach\ #1\ \algorithmicdo}

\newcommand\blfootnote[1]{
    \begingroup
    \renewcommand\thefootnote{}\footnote{#1}
    \addtocounter{footnote}{-1}
    \endgroup
}

\title{Smooth Ranking SVM via Cutting-Plane Method\cmmnt{for regularization}}

\author{
Erhan Can Ozcan$^1$
\and
Berk Görgülü$^2$\and
Mustafa G. Baydogan$^{3}$\And
Ioannis Ch. Paschalidis$^1$\\
\affiliations
$^1$Boston University\\
$^2$McMaster University \\
$^3$Bogazici University\\
\emails
\{cozcan, yannisp\}@bu.edu,
gorgulub@mcmaster.ca,
mustafa.baydogan@boun.edu.tr
}

\begin{document}
\maketitle

\begin{abstract}
The most popular classification algorithms are designed to maximize classification accuracy during training. However, this strategy may fail in the presence of class imbalance since it is possible to train models with high accuracy by overfitting to the majority class. On the other hand, the Area Under the Curve (AUC) is a widely used metric to compare classification performance of different algorithms when there is a class imbalance, and various approaches focusing on the direct optimization of this metric during training have been proposed. Among them, SVM-based formulations are especially popular as this formulation allows incorporating different regularization strategies easily. In this work, we develop a prototype learning approach that relies on cutting-plane method, similar to Ranking SVM, to maximize AUC. Our algorithm learns simpler models by iteratively introducing cutting planes, thus overfitting is prevented in an unconventional way. Furthermore, it penalizes the changes in the weights at each iteration to avoid large jumps that might be observed in the test performance, thus facilitating a smooth learning process. Based on the experiments conducted on 73 binary classification datasets, our method yields the best test AUC in 25 datasets among its relevant competitors.
\end{abstract}

\blfootnote{Under Review.}

\section{Introduction}

Classification models are frequently deployed in various real-world problems ranging from e-mail spam filters to medical diagnostic tests. Since the main goal in classification is correctly categorizing data into predetermined classes, accuracy is widely used as a target measure for evaluating the performance of different algorithms. However, class imbalance is a common problem that exists in most of the real-world datasets, and it can significantly affect the performance of classifiers that maximize accuracy during training. Moreover, in case of class imbalance, accuracy may be a misleading metric to assess the performance of a classifier \cite{rakotomamonjy2004optimizing}. For example, in a scenario where only 7\% of the emails are spam, the model can easily achieve 93\% accuracy without predicting any email as spam, although such a model has no practical usefulness.

Overcoming class imbalance is essential for the success of classification algorithms, and various techniques have been proposed to handle imbalanced datasets \cite{leevy2018survey}. While some works tackle this problem at the data-level by using over-sampling and under-sampling methods \cite{khoshgoftaar2007learning,van2007experimental}, cost sensitive algorithm-level alternatives that change the loss function during the training process have gained popularity recently \cite{cao2013optimized,wang2012using}. However, cost sensitive learning solutions rely on explicit knowledge of the misclassification costs, which can be unknown in most of the cases \cite{ali2013classification}. Therefore, the use of these solutions can be limited in real-world problems.


On the other hand, approaches maximizing the Area Under the Receiver Operating Characteristics (ROC) Curve (AUC) during training may be better algorithm-level alternatives to consider when there is a class imbalance within a dataset \cmmnt{to mitigate the class imbalance problem} \cite{ling2003auc}. AUC concentrates on the relative prediction scores of samples \cite{bradley1997use}, and quantifies the ranking capability of a model based on pairwise comparisons. Although it is primarily a ranking related metric, its applications are extensively used in classification problems as well\cmmnt{in addition to ranking problems} \cite{calders2007efficient,norton2019maximization}. Furthermore, due to its robustness to class imbalance, AUC is an important criterion to assess the performance of a model, and it is preferred over classification accuracy \cite{ling2003auc}.

Unfortunately, AUC is neither a continuous nor a differentiable function \cite{yan2003optimizing}, which makes it difficult to optimize. In \cite{chang2012integer}, a mixed integer programming problem is proposed to maximize the AUC exactly. In theory, solving this problem yields the highest possible training AUC. However, in this formulation, the number of binary variables grows quadratically due to pairwise comparisons. Therefore, it takes significant amount of time to optimize these models in practice. Furthermore, it can lead to overfitting and poor generalization. After relaxing the integrality constraints, AUC can be approximated via a piecewise linear function. Based on this idea, variuous AUC maximizing approaches such as Ranking SVM \cite{joachims2002optimizing}, RankBoost \cite{freund2003efficient}, and RankNet \cite{burges2005learning} have been developed. In this sense, the problem formulation we propose in this study has similarities with SVM-based models focusing on AUC maximization.

Most real-world problems exhibit non-linear relations, and the linear models cannot effectively represent these complex patterns. Kernel functions can be helpful to learn complex decision boundaries, and the radial basis function (RBF) kernel is one of the popular choices. This function provides a proximity measure by scaling the Euclidean distance between data points via a shape parameter, which can affect the performance of the models substantially \cite{fasshauer2007choosing}. Therefore, it is essential to choose the right kernel for each problem via parameter tuning operations. On the other hand, the Euclidean distance calculation can be considered as a non-linear transformation of the original raw features, and utilizing the Euclidean distances between data points as features can help the model to learn non-linear relations \cite{duin2010feature}. Moreover, it is possible to extract features with high discriminatory power by learning prototype points residing on the original feature space \cite{alvarez2022fuzzy,zhang2015dissimilarity,biehl2013distance}. The study in \cite{ozcan2023column} leverages this idea, and devises an iterative algorithm that finds the prototype points by solving a subproblem at each iteration. In this study, we build our algorithm based on this prototype learning strategy.

One common problem of learning methods is that they can easily overfit the training data, and they fail to generalize well to new data points \cite{chen2020distributionally}. This issue deepens even more when the learned model is too complex. In order to prevent overfitting, a number of techniques have been developed. State-of-the-art approaches either incorporate a regularization term to the optimization problem \cite{ghaddar2018high,jimenez2021novel} or apply different feature selection techniques to the data prior to optimization \cite{li2017feature}. The first idea is well explored in the context of Ranking SVMs, and some effective variants such as Ranking-SVM with the $L_2$-norm regularizer \cite{rakotomamonjy2004optimizing,joachims2002optimizing} and Ranking-SVM with the $L_1$-norm regularizer \cite{ataman2005optimizing} are proposed. These approaches prevent coefficients from growing too much by penalizing their $L_2$-norm and $L_1$-norm in the objective function, respectively, and the impact of the regularization term on the objective function is controlled via a constant cost parameter, $C$. However, as the choice of the cost parameter, $C$, can dramatically affect the test performance of the learned models, it is required to conduct cross-validations during the experiments to select the best $C$. Another study in \cite{ming2019robust} suggests combining $L_1$ and $L_2$-norm in the SVM context for better regularization and classification performance. Furthermore, the study in \cite{bertsimas2020prescriptive} proposes a sparse SVM classifier by partitioning classes into clusters. On the other hand, feature selection refers to identifying the most relevant features to represent the patterns in data before model learning takes place. However, since there is no general recipe to follow during this process, feature selection usually requires a solid domain knowledge. One recent study in \cite{ozcan2023column} introduces the Ranking-CG Prototype algorithm that employs column generation (CG) on a variant of Ranking SVM as an internal feature selection mechanism, and shows that column generation can be an important tool to prevent overfitting. \cmmnt{However, the formulation proposed in \cite{ozcan2023column} learns a model by bounding the $L_{\infty}$-norm of the weights, which can result in learning suboptimal models in some cases, thus degrading the overall performance. In our study, we propose a new formulation to overcome this issue. Moreover, in Ranking-CG Prototype algorithm, column generation iterations are ceased with the help of a convergence parameter, which affects the test performance of the model. Therefore, the value of this parameter is determined via cross-validation. The proposed algorithm allows us to}\cmmnt{Although there are numerous studies \cite{ataman2005optimizing,dedieu2019solving,zhang2013analysis} that employ the column generation idea in SVM-based models to accelerate the optimization, the use of column generation to prevent overfitting is recent, and it can be explored further.}Although there are numerous studies \cite{ataman2005optimizing,dedieu2019solving,zhang2013analysis} that employ the column generation idea in SVM-based models to accelerate the optimization, the benefit of using of column generation to prevent overfitting has only recently being considered. Therefore, exploring this idea further is promising.

In this study, we propose Smooth Ranking-CG Prototype to maximize AUC. Similar to Ranking-CG Prototype, this method generates cutting-planes based on column generation to learn prototype points. However, the Ranking-CG Prototype learns a model by bounding the $L_{\infty}$-norm of the weights \cite{ozcan2023column}, which can result in learning sub-optimal models in some cases, thus degrading the overall performance. \cmmnt{In our study, we propose a new formulation to overcome this issue.}In our study, we propose a formulation, which does not restrict the $L_{\infty}$-norm of the weights. Therefore, this is the first work where the regularization is achieved via only column generation as opposed to Ranking-CG Prototype. Moreover, in Ranking-CG Prototype algorithm, column generation iterations are ceased with the help of a convergence parameter, which affects the test performance of the model. Therefore, the value of this parameter is determined via cross-validation. On the other hand, in our formulation, the convergence parameter plays a less critical role, which helps us to remove the convergence parameter from our algorithm. \cmmnt{According to the experiments conducted on 74 publicly available datasets, \red{the Smooth Ranking-CG Prototype provides the highest AUC in 22 datasets. Furthermore, it is possible to reduce the number of features significantly by using the proposed approach.}}Finally, by conducting experiments over 73 publicly available datasets, we show that the Smooth Ranking-CG Prototype can be a competitive approach since it outperforms all relevant competitors by providing the highest AUC in 25 datasets.

\section{Notations and Background}

\cmmnt{This study presents a prototype learning strategy to maximize AUC similar to \cite{ozcan2023column}. Therefore, we recall the relevant concepts and the procedures by following their notation mostly. }Given a labeled dataset $\mathcal{Z}=(\mathcal{X},\mathbf{y}$), let $(x_i,y_i)$ be the $i^\text{th}$ (feature, label) pair where $x_i \in \mathbb{R}^{d}$ denotes a $d$-dimensional feature vector and $y_i \in \{-1,1\}$ denotes the class label\cmmnt{for the $i^\text{th}$ data point}. \cmmnt{Let $(x_i,y_i)$ be a (feature, label) pair where $x_i \in \mathbb{R}^{d}$ denotes a $d$-dimensional feature vector and $y_i \in \{-1,1\}$ denotes the class label. Also let $(\mathcal{X},\mathbf{y}$) be the collection of all pairs in the data.}Suppose our goal is to learn a score function $\phi(\cdot)$ that maximizes the AUC statistic represented as follows:
\begin{align}
\text{AUC}_{\phi}=\small \frac{\sum\limits_{p\epsilon \mathcal{P}}\sum\limits_{n\epsilon \mathcal{N}} { \mathbbm{1}{\{\phi(x_p) > \phi(x_n)\}}} }{|\mathcal{P}|\ |\mathcal{N}|},\quad\label{auc_def}
\end{align}
\noindent where $\mathcal{P}$ and $\mathcal{N}$ respectively contain the indexes of positive and negative instances, {$\mathbbm{1}\{\cdot\}$} is the indicator function, and $|.|$ denotes the cardinality of a set. Due to the counting operator involved in AUC, this problem is non-convex regardless of the form of the chosen score function $\phi(\cdot)$. Therefore, the direct optimization of this metric can be computationally intractable for large datasets.\cmmnt{Since AUC involves a counting operator, the direct optimization of this metric requires solving a mixed-integer programming problem, which can be computationally intractable for large datasets.}

However, a good approximation can be obtained by replacing the indicator function with the following hinge loss function:
\[\max\left(0\ ,\ 1-\left(\phi(x_p)-\phi(x_n)\right)\right).\]
If the score function $\phi(\cdot)$ is selected as an affine function, e.g., $\phi(x)=w^Tx+b$, a convex ranking approach with the $L_2$-norm regularizer on the parameter vector $w$ can be formulated as follows:

\begin{align}
\label{rank_hinge}
\min_{w}\quad  \frac{1}{2}w^{T}w + C\sum\limits_{p\epsilon \mathcal{P}}\sum\limits_{n\epsilon \mathcal{N}}\max\left(0\ ,\ 1-\left(w^T(x_p-x_n)\right)\right),
\end{align}
where $C\geq0$ is the regularization penalty. \cmmnt{Finally, by using classic convex optimization techniques}We define $\xi_{p,n}$ as the slack variable that measures the margin of error while comparing a positive instance with index $p \in \mathcal{P}$ with a negative instance with index $n \in \mathcal{N}$. By upper bounding the term in the summation by $\xi_{p,n}$, Problem \eqref{rank_hinge} can be transformed into the Ranking SVM algorithm in \cite{joachims2002optimizing}:

\begin{subequations}
\begin{alignat}{3}
&\large \underset{w,\xi} \min\ \quad &&\frac{1}{2}w^{T}w + C\sum\limits_{p\epsilon \mathcal{P}}\sum\limits_{n\epsilon \mathcal{N}}{\xi_{p,n}} && \label{eq:4}\\
&\text{ s.t.}\quad && w^{T}(x_p-x_n) \geq 1- \xi_{p,n}, &&\quad\forall p\in \mathcal{P},\forall n \in \mathcal{N},\label{eq:5}\\
&\quad &&\xi_{p,n} \geq0,  &&\quad\forall p\in \mathcal{P},\forall n \in \mathcal{N},
\end{alignat}
\end{subequations}
\cmmnt{where $\xi_{p,n}$ is the slack variable that measures the margin of error while comparing a positive instance with index $p \in \mathcal{P}$ with a negative instance with index $n \in \mathcal{N}$. }Note that Ranking SVM focuses on pairwise comparisons of the positive and negative instances and minimizes the total error in these pairwise comparisons. Therefore, it can learn classifiers that are robust to class imbalance problem.\cmmnt{Maximizing AUC instead of classification accuracy addresses the issues caused by the class imbalance in the data by considering the pairwise comparisons of the positive and negative instances which implicitly balances the class weights.} However, problems such as overfitting and the inability to model non-linear relations persist in this formulation.

To model non-linear relations, the study in \cite{ozcan2023column} proposes learning a score function based on the Euclidean distance to some prototype points:

\begin{equation}
\label{score_func}
\phi(x) := \phi(x, w, \mathcal{Q}) = \sum\limits_{t=1}^{|\mathcal{Q}|}w^{(t)}\lVert x-q_t \rVert,
\end{equation}
where $x$ denotes the feature vector of an instance, $w^{(t)}$ is the $t^{th}$ element of the weight vector $w$, $\mathcal{Q}$ is the set of $d-$dimensional prototype points, and $q_t$ denotes a single a point in $\mathcal{Q}$, i.e., $q_t \in \mathcal{Q}$, $\forall t=\{1, …, |\mathcal{Q}|\}$. Accordingly, for any given $\mathcal{Q}$, they find the weight vector $w$ by solving the following problem $\mathcal{F}(\mathcal{Q})$.

\vspace{2mm}
\underline{{$\mathcal{F}(\mathcal{Q}):$}}
\begin{subequations}\label{prob_2}
\begin{align}
&\underset{w,\xi}\min\ &&\sum\limits_{p\in \mathcal{P}}\sum\limits_{n\in \mathcal{N}} {\xi_{p,n}} \label{eq:non_obj}\\
\notag&\text{ s.t.}\ &&
\sum\limits_{t=1}^{|\mathcal{Q}|}w^{(t)}(\lVert x_p-q_t \rVert  -\lVert x_n-q_t \rVert) \geq 1- \xi_{p,n},\\ 
\label{eq_soft}&\ &&\ \ \ \quad\qquad\qquad\qquad\qquad\forall p\in \mathcal{P},\forall n \in \mathcal{N},\\
\label{eq_bound}&\ &&-1\leq w^{(t)} \leq 1,\ \ \ \qquad\forall t = 1, \dots,|\mathcal{Q}|, \\
&\quad &&\xi_{p,n} \geq0,  \qquad\qquad\qquad\forall p\in \mathcal{P},\forall n \in \mathcal{N}. \label{eq:non_obj_end}
\end{align}
\end{subequations}


As the number of points in the set $\mathcal{Q}$ increases, the learned model becomes more complex. Therefore, the learned model is likely to overfit when the set $\mathcal{Q}$ includes many points. However, the optimization problem given in \eqref{prob_2} is a linear programming problem, and it is possible to design a column generation strategy, a popular technique relying on duality \cite{bertsimas1997introduction}, for finding a new prototype point to add to the set $\mathcal{Q}$. Therefore, column generation can be an effective tool to control the complexity of the model.

\section{Proposed Approach}

The problem given in \eqref{prob_2} can be seen as a variant of Ranking-SVM with the $L_\infty$-norm regularizer since constraint \eqref{eq_bound} restricts the weights between $-1$ and $1$. However, bounding the weight vector into a certain region may result in sub-optimal solutions, thus poor performance in practice. Hence, it is important to design a formulation that does not include this constraint. 

The Simplex algorithm, which is frequently used to solve linear programming problems, searches for the optimal solution by jumping from one vertex to an adjacent vertex \cite{bertsimas1997introduction}. \cmmnt{Therefore, the optimal solution may change drastically from iteration to iteration, which may cause a fluctuation in test AUC.}Therefore, the optimal solution may change drastically from iteration to iteration. In our context, this is undesirable as the large changes in the weight vector may result in fluctuations in test AUC. Therefore, removing constraint \eqref{eq_bound} requires a modification to the optimization problem solved at each iteration.

Since our goal is to prevent weights from changing dramatically from iteration to iteration, we introduce an additional term, which penalizes the change in weights via a non-negative smoothing parameter $C$, to the objective function of the problem in \eqref{prob_2}. Suppose that $w_{\text{old}}$ denotes the weight vector learned before adding a new prototype point to the set $\mathcal{Q}$, then, for a given $\mathcal{Q}$, $w_{\text{old}}\ \in\ \mathbb{R}^{|\mathcal{Q}|-1}$, and with a smoothing parameter, $C$, we formulate the problem as follows:
\begin{subequations}\label{prob_3}
\begin{align}
&\underset{w,\xi}\min &&\sum\limits_{p\in \mathcal{P}}\sum\limits_{n\in \mathcal{N}} {\xi_{p,n}} + C \sum_{t=1}^{|\mathcal{Q}|-1} |w^{(t)} - w_{\text{old}}^{(t)}| \label{eqnew:non_obj}\\
\notag&\text{ s.t.} &&
\sum\limits_{t=1}^{|\mathcal{Q}|}w^{(t)}(\lVert x_p-q_t \rVert  -\lVert x_n-q_t \rVert) \geq 1- \xi_{p,n},\\ \label{eqnew_soft}&\ &&\qquad\qquad\qquad\qquad\qquad\forall p\in \mathcal{P},\forall n \in \mathcal{N},\\
&\quad &&\xi_{p,n} \geq0,  \qquad\qquad\qquad\forall p\in \mathcal{P},\forall n \in \mathcal{N}. \label{eqnew:non_obj_end}
\end{align}
\end{subequations}

By following standard reformulation procedures for linear programs, we can replace the absolute values in the objective function \eqref{eqnew:non_obj} by an auxiliary vector $s$, and obtain an equivalent formulation to the problem in \eqref{prob_3}, which we denote as $\mathcal{F}(\mathcal{Q},w_{\text{old}},C)$.

\vspace{2mm}
\underline{$\mathcal{F}(\mathcal{Q},w_{\text{old}},C)$:}
\begin{subequations}\label{prob_4}
\begin{align}
&\underset{w,\xi,s}\min &&\sum\limits_{p\in \mathcal{P}}\sum\limits_{n\in \mathcal{N}} {\xi_{p,n}} + C \sum_{t=1}^{|\mathcal{Q}|-1} s^{(t)} \label{eqnew:no_abs}\\
\notag&\text{ s.t.} &&
\sum\limits_{t=1}^{|\mathcal{Q}|}w^{(t)}(\lVert x_p-q_t \rVert  -\lVert x_n-q_t \rVert) \geq 1- \xi_{p,n}, \\\label{eqnew_soft_conv}&\ &&\qquad\qquad\qquad\qquad\qquad\forall p\in \mathcal{P},\forall n \in \mathcal{N},\\
\label{eqnew_abs1}&\ &&s^{(t)} - w^{(t)} \geq - w_{\text{old}}^{(t)}, \ \ \quad\forall t = 1, \dots,|\mathcal{Q}|-1,\\
\label{eqnew_abs2}&\ &&s^{(t)} + w^{(t)} \geq w_{\text{old}}^{(t)}, \ \qquad\forall t = 1, \dots,|\mathcal{Q}|-1,\\
&\ &&s^{(t)} \geq 0, \ \ \qquad\qquad\qquad\forall t = 1, \dots,|\mathcal{Q}|-1,\\
&\ &&\xi_{p,n} \geq0,  \ \ \qquad\qquad\qquad\forall p\in \mathcal{P},\forall n \in \mathcal{N}. \label{eqnew:non_obj_end_2}
\end{align}
\end{subequations}

Column generation in linear programming is built upon duality theory, and adding a new column to the primal problem results in adding a new cutting-plane to the dual problem \cite{bertsimas1997introduction}. Hence, we can design a mechanism to find new prototype point to add to the set $\mathcal{Q}$ by solving a sub-problem that is formulated based on the information provided by dual variables. Suppose that the problem in \eqref{prob_4} is formulated for a given $\mathcal{Q}$, $w_{\text{old}}$, and $C$, then the corresponding dual problem is as follows:
\cmmnt{Column generation in linear programming is built upon duality theory, and new prototype point to add to the set $\mathcal{Q}$ is found by solving a sub-problem that is formulated based on the information provided by dual variables. Suppose that the problem in \eqref{prob_4} is formulated for a given $\mathcal{Q}$, $w_{\text{old}}$, and $C$, then the corresponding dual problem is as follows:}
\begin{subequations}\label{dual:lp_main}
\begin{align}\label{dual:lp}
&\underset{\pi, \alpha, \beta}\max &&\sum\limits_{p\in \mathcal{P}}\sum\limits_{n\in \mathcal{N}} {\pi_{p,n}}-\sum\limits_{t=1}^{|\mathcal{Q}|}{(\alpha^{ (t)}-\beta^{(t)})}w_{\text{old}}^{(t)}\\
&\text{ s.t.} &&
\notag\sum\limits_{p\in \mathcal{P}}\sum\limits_{n\in \mathcal{N}} {\pi_{p,n}(\lVert x_p-q_t \rVert  -\lVert x_n-q_t \rVert)} \\&\ &&\quad-{\alpha^{ (t)}}+{\beta^{ (t)}}=0, \quad t=1,2,...,|\mathcal{Q}|-1,\label{dual:cons1}\\
\notag&\ && \sum\limits_{p\in \mathcal{P}}\sum\limits_{n\in \mathcal{N}} {\pi_{p,n}(\lVert x_p-q_t \rVert  -\lVert x_n-q_t \rVert)}=0, \\&\ && \qquad\qquad\qquad\qquad\qquad t=|\mathcal{Q}|,\label{dual:cons2}\\
&\ &&\alpha^{ (t)} + \beta^{ (t)} \leq C, \quad\qquad t=1,2,...,|\mathcal{Q}|-1,\label{dual:cons3}\\
&\ &&\pi_{p,n} \leq1, \ \ \ \quad\qquad\qquad\forall p\in \mathcal{P},\forall n \in \mathcal{N},\label{dual:cons4}\\
&\ &&\pi_{p,n} \geq0, \ \ \ \quad\qquad\qquad\forall p\in \mathcal{P},\forall n \in \mathcal{N},\label{dual:cons5}\\
&\ &&{\alpha^{ (t)}} \geq0, \qquad\qquad\qquad t=1,2,...,|\mathcal{Q}|-1,\label{dual:cons6}\\
&\ &&{\beta^{ (t)}} \geq0, \qquad\qquad\qquad t=1,2,...,|\mathcal{Q}|-1\label{dual:cons7},
\end{align}
\end{subequations}
where $\pi_{p,n}$ for $p \in \mathcal{P}, n \in \mathcal{N}$ are the dual variables associated with the \textit{soft-margin} constraints given in \eqref{eqnew_soft_conv}, and $\alpha:=(\alpha^{(1)},\alpha^{(2)},\dots,\alpha^{(|\mathcal{Q}|-1)})$, $\beta:=(\beta^{(1)},\beta^{(2)},\dots,\beta^{(|\mathcal{Q}|-1)})$ are the dual variables associated with the constraints given in \eqref{eqnew_abs1} and \eqref{eqnew_abs2}, respectively. According to the duality theory, it is enough to optimize the primal problem to retrieve the optimal dual solution or vice versa. Moreover, adding a new prototype point to the primal problem results in generating the constraint in \eqref{dual:cons2}. Therefore, given the optimal solution of Problem \eqref{dual:lp_main}, denoted by $\pi^*$, if there exists a $q\ \in\ \mathbb{R}^d$ such that
\begin{equation}
\label{eq_dual_feas}
\sum\limits_{p\in \mathcal{P}}\sum\limits_{n\in \mathcal{N}} {\pi^*_{p,n}(\lVert x_p-q \rVert  -\lVert x_n-q \rVert)}  \neq 0,
\end{equation}
then adding $q$ to the set $\mathcal{Q}$ can provide an improvement to the model as it harms the dual feasibility. In order to find new prototype point, $\tilde{q}$, we solve the following non-linear sub-problem by using a gradient based optimization approach:  

\begin{align}\label{eq:two}
\tilde{q}=\large \underset{q \in \mathbb{R}^d} \argmax \left|\sum\limits_{p\epsilon P}\sum\limits_{n\epsilon N} {\pi_{p,n}^*(\lVert x_p-q \rVert  -\lVert x_n-q \rVert)}\right|. 
\end{align}

Then, the pseudo-code of Smooth Ranking-CG Prototype is provided in Algorithm \ref{iter:three}.

\begin{algorithm}[ht]
\caption{Smooth Ranking-CG Prototype}
\label{iter:three}
\noindent \textbf{Input:} Smoothing parameter ($C>0$), Set of all points ($\mathcal{X}$)\\
\noindent \textbf{Output:} Set of prototype points ($\mathcal{Q}$), Optimal weights ($w^*$)
\begin{algorithmic}[1]
\State Initialize $t=1$, $z_0=10^{-6}$, $\mathcal{Q}=\emptyset$
\State Randomly select a point $\Tilde{q} \in \mathcal{X}$
\Repeat 
\State $\mathcal{Q} = \mathcal{Q} \cup \Tilde{q}$
\State Solve $\mathcal{F}(\mathcal{Q},w_{\text{old}},C)$ in \eqref{prob_4} to obtain $w^*$ and $\pi^*$
\State Define $\phi(.):=\phi(\ .\ ,w^*,\mathcal{Q})$, and set $z_t=\text{AUC}_{\phi}$
\State Find $\tilde{q}=\large \underset{q \in \mathbb{R}^d} \argmax \left|\sum\limits_{p\epsilon P}\sum\limits_{n\epsilon N} {\pi_{p,n}^*(\lVert x_p-q \rVert  -\lVert x_n-q \rVert)}\right| $
\State Set $w_{\text{old}} = w^*$
\State $t=t+1$
\Until{$\ \ \frac{\left|z_{t-1}\ - \ z_{t}\right|}{z_{t-1}} < 0.01 $} 
\end{algorithmic}
\end{algorithm}

The iterations in Algorithm \ref{iter:three} are initiated by adding a randomly selected point $\tilde{q}$ from the feature space to the set $\mathcal{Q}$. After solving the problem $\mathcal{F}(\mathcal{Q},w_{\text{old}},C)$ and obtaining $w^*$ and $\pi^*$, we solve the non-linear sub-problem \eqref{eq:two} to find the prototype point $\tilde{q}$. Finally, iterations continue until the algorithm converges.

Unfortunately, \eqref{eq:two} is a complex function that is neither convex nor concave. Hence, it might have many local optima and the gradient based algorithms may get stuck at one of these points. Techniques such as using adaptive learning rates and introducing a momentum term are proved to be helpful to avoid local optimums during the global optimum search in practice, but it is not possible to guarantee that the algorithm reaches the global optimum. However, by using a proper initialization strategy, the gradient descent based algorithm may converge to better local optima, and we can utilize available training data to design this initialization strategy. Suppose we have the dual solution, $\pi^*$, then the optimization of the problem in \eqref{eq:two} can be initiated starting from $\hat{q} \in \mathcal{X}$:

\begin{align}
    \hat{q} = \underset{q \in \mathcal{X}}\argmax \left|\sum\limits_{p\in \mathcal{P}}\sum\limits_{n\in \mathcal{N}} {\pi^{*}_{p,n}(\lVert x_p-q \rVert - \lVert x_n-q \rVert)}\right|. \label{eq:sub_problem_}
\end{align}

Notice that the problem in \eqref{eq:sub_problem_} is easy to solve since we have finitely many training samples. After finding $\hat{q}$, the optimization of \eqref{eq:two} continues for a specific number of iterations unless convergence is observed. The details of the gradient based algorithm we employ to solve \eqref{eq:two} are provided in Section \ref{exps}.

\section{Experiments}\label{exps}

%

We conduct different set of experiments to show the benefits of the proposed approach. Section \ref{sec:experimental_setup} provides the experimental setup; \cmmnt{Section \ref{sec:prototype_comparison} demonstrates that smooth Ranking-CG Prototype can address the fluctuating test AUC issue that Ranking-CG Prototype may exhibit if the weight vector is unbounded.}Section \ref{sec:prototype_comparison} demonstrates that the smooth Ranking-CG Prototype can be helpful to prevent the fluctuations in test AUC at consecutive iterations, which Ranking-CG Prototype may exhibit if the weight vector is unbounded; and Section \ref{sec:classification} shows the classification performance of the proposed approach with respect to different benchmarks.

\subsection{Experimental Setup}\label{sec:experimental_setup}

The model is implemented in the Python programming language, and the code is publicly available\footnote{\url{https://github.com/erhancanozcan/SmoothRankingSVM}.}. While Gurobi 9.5.1 is utilized to solve linear programs, the Adam optimizer in TensorFlow 2.10.0 is utilized to solve non-linear sub-problem. We test the performance of the proposed method on binary classification problems that are publicly available on the UCI \cite{Dua:2019} and KEEL \cite{alcala2011keel} repositories. The chosen datasets vary in terms of the number of features, the number of instances, and the severity of class imbalance, and the dataset characteristics are summarized in Appendix A.



The model proposed in this study is relevant to Ranking SVM-based models, thus we consider three Ranking SVM variants as benchmark algorithms. We refer to Ranking SVM with $L_2$, $L_1$ and $L_{\infty}$ regularizers as Ranking SVM, $L_1$ Ranking, and $L_{\infty}$ Ranking, respectively. While we utilize the Ranking SVM implementation in the dlib package  \cite{dlib09}, we implement other versions of the Ranking SVM, and solve them via Gurobi. To guarantee fair comparisons among models, all the aforementioned methods are trained on the pairwise dissimilarity matrix obtained based on Euclidean distances instead of the raw features. Therefore, the upper limit for the number of features that any method can consider is equal to the number of points in the training data, except for the prototype learning approaches. Lastly, we include the Ranking-CG Prototype method that achieves regularization internally via column generation as described in \cite{ozcan2023column} among the benchmarks we consider.

The performance of the benchmark methods heavily depends on the value of its parameter, which controls the model complexity, and the best value of this parameter is chosen via 5-fold cross-validation for each dataset during the experiments. To ensure a fair comparison, the number of options considered for the complexity parameter is equal across different approaches, and the details of those parameters are available in Appendix B.

On the other hand, the Smooth Ranking-CG Prototype has a smoothing parameter, $C$, which penalizes the changes in weights at each iteration. This parameter is selected based on a 5-fold cross-validation from thirteen values in the interval of $[10^{-6},1.0]$ (see Appendix B for the details). Additionally, during the optimization of the sub-problem in Equation \eqref{eq:two}, we employ the Adam optimizer with its default parameters, and the optimization of the sub-problem continues as long as the change in objective ratio is greater than $10^{-5}$. If this convergence criteria is not met in $1000$ iterations, the optimization of the sub-problem is interrupted, and the algorithm continues the iterations by adding the best prototype point to the set $\mathcal{Q}$.

\subsection{Stabilizing the Progress of Test AUC}\label{sec:prototype_comparison}

As mentioned earlier, the Ranking-CG Prototype learns a model by solving the problem given in \eqref{prob_2}, thus the model weights are restricted between $-1$ and $1$. Due to this bound, the learned model can be sub-optimal in some cases, thus showing poor performance. However, removing this bound can be devastating for the test performance of the model as it will make the model less predictable. The smooth Ranking-CG Prototype addresses this issue by penalizing the change of the weight vector instead of bounding the $L_{\infty}$ norm of the weight vector as opposed to the Ranking-CG Prototype.

\begin{figure*}[!h]%
    \centering
    \subfloat[Scatter Plot of XOR Dataset]{{\includegraphics[width=.4\textwidth]{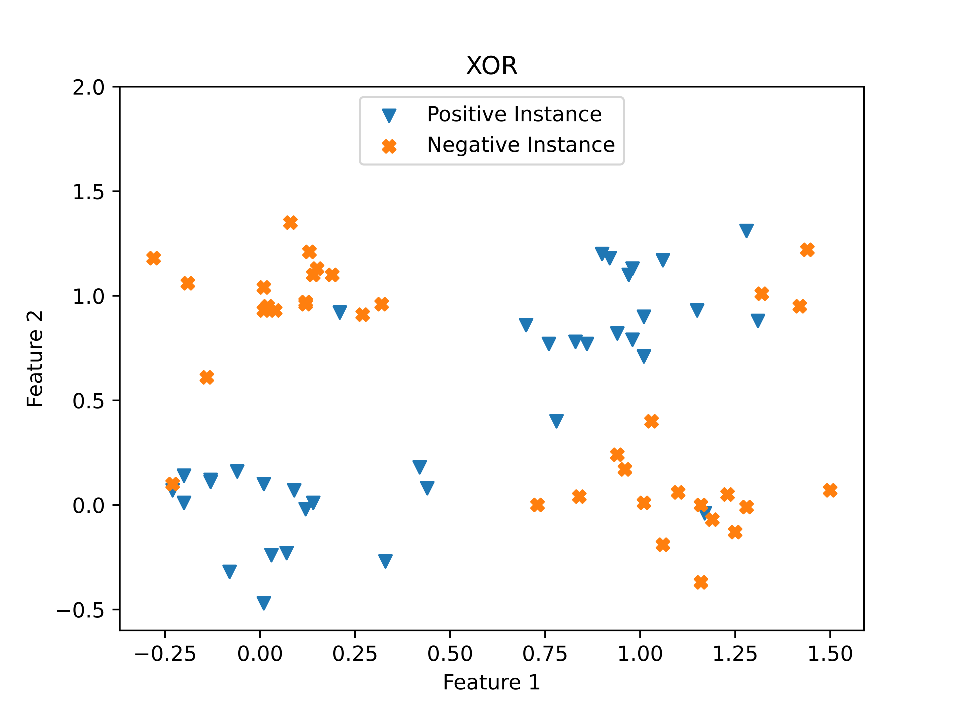} }\label{fig:xor}}
    \subfloat[Test AUC]{{\includegraphics[width=.4\textwidth]{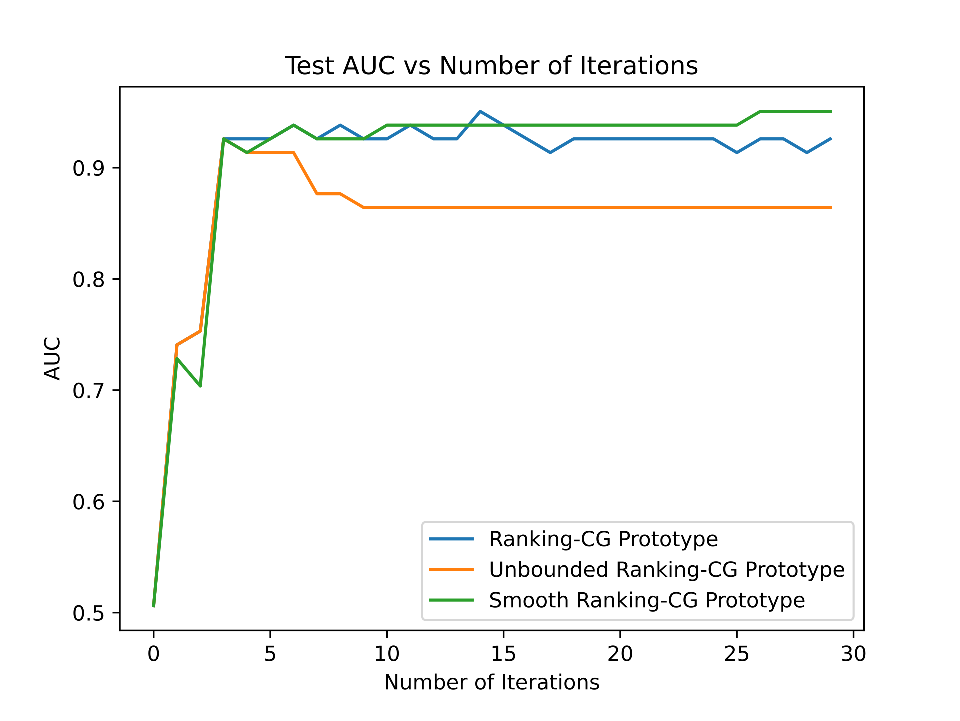} }\label{fig:testAUC}}
    \caption{Changes in the Test AUC with the number of iterations (each newly added point to the set $\mathcal{Q}$) for the Ranking-CG Prototype, Unbounded Ranking-CG Prototype and Smooth Ranking-CG Prototype methods.}%
    \label{fig:one}%
\end{figure*}

In order to investigate the case described above, we implement a version of Ranking-CG Prototype by removing the bound on the weight vector, which we refer as Unbounded Ranking-CG Prototype; and we monitor the change in the test AUC values after adding a new point to the set $\mathcal{Q}$ at each iteration for different approaches on a simple XOR problem with some noise. Figure \ref{fig:one} shows the scatter plot of the experimental XOR problem we consider, and compares the test AUC performance of our algorithm with the Ranking-CG Prototype and the Unbounded Ranking-CG Prototype as a function of the number of iterations.

According to Figure \ref{fig:testAUC}, the Unbounded Ranking-CG Prototype can be still vulnerable to overfitting unless the weight vector is bounded. Moreover, due to the fluctuations observed in the Ranking-CG Prototype and its unbounded version, the convergence parameter of these algorithms must to be tuned via cross-validation for a reliable performance. However, the performance of the Smooth Ranking-CG Prototype is more robust across the iterations since it allows the learned weights to change if and only if the improvement in the objective is large enough. Therefore, the convergence parameter in our approach is not as important as it is in the Ranking-CG, which allows us to remove this parameter from our algorithm.

\subsection{Classification Performance}\label{sec:classification}

We compare the Smooth Ranking-CG Prototype with the aforementioned relevant competitors by conducting experiments over 73 datasets, and the detailed results are available in Appendix C. According to these results, Table \ref{table:one} shows the number of datasets in which each approach provides the highest test AUC among all competitors (the first row), and the average $\%$ of features with coefficients' magnitude larger than $0.001$. 


\begin{table*}[!h]
\centering
\scalebox{.95}{
\begin{tabular}{lccccc}
\hline
\textbf{\begin{tabular}[c]{@{}c@{}}\end{tabular}} & \textbf{\begin{tabular}[c]{@{}c@{}}Smooth Ranking-CG \\ Prototype\end{tabular}} & \textbf{\begin{tabular}[c]{@{}c@{}}Ranking-CG \\ Prototype\end{tabular}} & \textbf{\begin{tabular}[c]{@{}c@{}}$L_1$ Ranking\end{tabular}} & \textbf{\begin{tabular}[c]{@{}c@{}}$L_\infty$ Ranking\end{tabular}} & \textbf{\begin{tabular}[c]{@{}c@{}}Ranking \\ SVM\end{tabular}} \\ \hline
   \# of Datasets with Highest AUC & 25                  & 26 & 30  & 24 & 24 \\
   \% of Features Used & 2\%                  & 7\% & 9\%  & 100\% & 73\%\\ \hline
\end{tabular}}
\caption{The number of datasets in which each method yields the highest AUC and the average of the \% of features used in 73 datasets. The sum of the first row is greater than 73 since more than one approach may yield the highest AUC in some datasets.}\label{table:one}
\end{table*}

The Smooth Ranking-CG Prototype yields the highest test AUC in 25 out of 73 datasets, which makes it as competitive as its relevant counterparts. Furthermore, as shown in Table \ref{table:one}, it utilizes significantly fewer features, thus it can be a useful approach when it is required to learn simpler models. Finally, to indicate the benefit of the proposed approach, we evaluate the performance of the proposed algorithm only in 47 datasets where the Ranking-CG Prototype cannot provide the highest test AUC, and the performance of algorithms over this subset of datasets is provided in Table \ref{table:two}.

\begin{table*}[!h]
\centering
\scalebox{.95}{
\begin{tabular}{lccccc}
\hline
\textbf{\begin{tabular}[c]{@{}c@{}}\end{tabular}} & \textbf{\begin{tabular}[c]{@{}c@{}}Smooth Ranking-CG \\ Prototype\end{tabular}} & \textbf{\begin{tabular}[c]{@{}c@{}}Ranking-CG \\ Prototype\end{tabular}} & \textbf{\begin{tabular}[c]{@{}c@{}}$L_1$ Ranking\end{tabular}} & \textbf{\begin{tabular}[c]{@{}c@{}}$L_\infty$ Ranking\end{tabular}} & \textbf{\begin{tabular}[c]{@{}c@{}}Ranking \\ SVM\end{tabular}} \\ \hline
   \# of Datasets with Highest AUC & 14                  & 0 & 18  & 12 & 14 \\ \hline
\end{tabular}}
\caption{The number of datasets in which each method yields the highest AUC in 47 datasets that the Ranking-CG Prototype fails to provide the highest test AUC. The sum of the first row is greater than 47 since more than one approach may yield the highest AUC in some datasets.}\label{table:two}
\end{table*}

\section{Conclusion}

This paper proposes a classifier maximizing AUC during training by incorporating a column generation-based prototype learning strategy via cutting-planes. Although a similar approach, Ranking-CG Prototype, has been proposed recently, we explain the problems that may arise in that algorithm, and attempt to handle those by modifying the optimization problem. The modification we propose allows us to remove the convergence parameter controlling the column generation iterations. Based on the extensive experiments conducted on 73 datasets, the Smooth Ranking-CG Prototype yields the highest test AUC in 25 datasets among its relevant competitors by learning simpler models.

We believe that the proposed method has two interesting aspects that can be explored further as future work. First, similar to Ranking-CG Prototype, the first prototype point is added to the set $\mathcal{Q}$ randomly to initiate the Smooth Ranking-CG Prototype, which may potentially have an impact on the performance of the algorithm. Second, the smoothing parameter, $C$, remains constant during the training of Smooth Ranking-CG Prototype. However, the role of this parameter is similar to learning rate in gradient-based algorithms, and setting the learning rate in an adaptive way is an important concept to improve the gradient-based algorithms. Inspired by this, one potential way to improve the performance of the Smooth Ranking-CG Prototype might be to design an algorithm adjusting the value of the smoothing parameter in an adaptive way at each iteration.

\section*{Acknowledgments}
Research was partially supported by the NSF under grants CCF-2200052, DMS-1664644, and IIS-1914792, by the DOE under grants DE-EE0009696 and DE-AR-0001282, and by the ONR under grant N00014-19-1-2571.

\bibliographystyle{named}
\bibliography{references}

\begin{thebibliography}{}

\bibitem[\protect\citeauthoryear{Alcal{\'a}-Fdez \bgroup \em et al.\egroup }{2011}]{alcala2011keel}
Jes{\'u}s Alcal{\'a}-Fdez, Alberto Fern{\'a}ndez, Juli{\'a}n Luengo, Joaqu{\'\i}n Derrac, Salvador Garc{\'\i}a, Luciano S{\'a}nchez, and Francisco Herrera.
\newblock Keel data-mining software tool: data set repository, integration of algorithms and experimental analysis framework.
\newblock {\em Journal of Multiple-Valued Logic \& Soft Computing}, 17, 2011.

\bibitem[\protect\citeauthoryear{Ali \bgroup \em et al.\egroup }{2013}]{ali2013classification}
Aida Ali, Siti~Mariyam Shamsuddin, and Anca~L Ralescu.
\newblock Classification with class imbalance problem.
\newblock {\em International Journal of Advances in Soft Computing and its Applications}, 5(3):176--204, 2013.

\bibitem[\protect\citeauthoryear{Alvarez \bgroup \em et al.\egroup }{2022}]{alvarez2022fuzzy}
Yanela~Rodr{\'\i}guez Alvarez, Mar{\'\i}a Matilde~Garc{\'\i}a Lorenzo, Yail{\'e}~Caballero Mota, Yaima~Filiberto Cabrera, Isabel M~Garc{\'\i}a Hilari{\'o}n, Daniela Machado~Montes de~Oca, and Rafael~Bello P{\'e}rez.
\newblock Fuzzy prototype selection-based classifiers for imbalanced data. case study.
\newblock {\em Pattern Recognition Letters}, 163:183--190, 2022.

\bibitem[\protect\citeauthoryear{Ataman and Street}{2005}]{ataman2005optimizing}
Kaan Ataman and W~Nick Street.
\newblock Optimizing area under the roc curve using ranking svms.
\newblock In {\em Proceedings of International Conference on Knowledge Discovery in Data Mining}, 2005.

\bibitem[\protect\citeauthoryear{Bertsimas and Tsitsiklis}{1997}]{bertsimas1997introduction}
Dimitris Bertsimas and John~N Tsitsiklis.
\newblock {\em Introduction to linear optimization}, volume~6.
\newblock Athena scientific Belmont, MA, 1997.

\bibitem[\protect\citeauthoryear{Bertsimas \bgroup \em et al.\egroup }{2020}]{bertsimas2020prescriptive}
Dimitris Bertsimas, Michael~Lingzhi Li, Ioannis~Ch Paschalidis, and Taiyao Wang.
\newblock Prescriptive analytics for reducing 30-day hospital readmissions after general surgery.
\newblock {\em PloS one}, 15(9):e0238118, 2020.

\bibitem[\protect\citeauthoryear{Biehl \bgroup \em et al.\egroup }{2013}]{biehl2013distance}
Michael Biehl, Barbara Hammer, and Thomas Villmann.
\newblock Distance measures for prototype based classification.
\newblock In {\em International Workshop on Brain-Inspired Computing}, pages 100--116. Springer, 2013.

\bibitem[\protect\citeauthoryear{Bradley}{1997}]{bradley1997use}
Andrew~P Bradley.
\newblock The use of the area under the {ROC} curve in the evaluation of machine learning algorithms.
\newblock {\em Pattern recognition}, 30(7):1145--1159, 1997.

\bibitem[\protect\citeauthoryear{Burges \bgroup \em et al.\egroup }{2005}]{burges2005learning}
Chris Burges, Tal Shaked, Erin Renshaw, Ari Lazier, Matt Deeds, Nicole Hamilton, and Greg Hullender.
\newblock Learning to rank using gradient descent.
\newblock In {\em Proceedings of the 22nd international conference on Machine learning}, pages 89--96, 2005.

\bibitem[\protect\citeauthoryear{Calders and Jaroszewicz}{2007}]{calders2007efficient}
Toon Calders and Szymon Jaroszewicz.
\newblock Efficient {AUC} optimization for classification.
\newblock In {\em European conference on principles of data mining and knowledge discovery}, pages 42--53. Springer, 2007.

\bibitem[\protect\citeauthoryear{Cao \bgroup \em et al.\egroup }{2013}]{cao2013optimized}
Peng Cao, Dazhe Zhao, and Osmar Zaiane.
\newblock An optimized cost-sensitive svm for imbalanced data learning.
\newblock In {\em Pacific-Asia conference on knowledge discovery and data mining}, pages 280--292. Springer, 2013.

\bibitem[\protect\citeauthoryear{Chang}{2012}]{chang2012integer}
Allison~An Chang.
\newblock {\em Integer optimization methods for machine learning}.
\newblock PhD thesis, Massachusetts Institute of Technology, 2012.

\bibitem[\protect\citeauthoryear{Chen \bgroup \em et al.\egroup }{2020}]{chen2020distributionally}
Ruidi Chen, Ioannis~Ch Paschalidis, et~al.
\newblock Distributionally robust learning.
\newblock {\em Foundations and Trends{\textregistered} in Optimization}, 4(1-2):1--243, 2020.

\bibitem[\protect\citeauthoryear{Dedieu and Mazumder}{2019}]{dedieu2019solving}
Antoine Dedieu and Rahul Mazumder.
\newblock Solving large-scale l1-regularized {SVM}s and cousins: the surprising effectiveness of column and constraint generation.
\newblock {\em arXiv preprint arXiv:1901.01585}, 2019.

\bibitem[\protect\citeauthoryear{Dua and Graff}{2017}]{Dua:2019}
Dheeru Dua and Casey Graff.
\newblock {UCI} machine learning repository, 2017.

\bibitem[\protect\citeauthoryear{Duin \bgroup \em et al.\egroup }{2010}]{duin2010feature}
Robert~PW Duin, Marco Loog, El{\.z}bieta P\c{e}kalska, and David~MJ Tax.
\newblock Feature-based dissimilarity space classification.
\newblock In {\em International Conference on Pattern Recognition}, pages 46--55. Springer, 2010.

\bibitem[\protect\citeauthoryear{Fasshauer and Zhang}{2007}]{fasshauer2007choosing}
Gregory~E Fasshauer and Jack~G Zhang.
\newblock On choosing “optimal” shape parameters for rbf approximation.
\newblock {\em Numerical Algorithms}, 45:345--368, 2007.

\bibitem[\protect\citeauthoryear{Freund \bgroup \em et al.\egroup }{2003}]{freund2003efficient}
Yoav Freund, Raj Iyer, Robert~E Schapire, and Yoram Singer.
\newblock An efficient boosting algorithm for combining preferences.
\newblock {\em Journal of machine learning research}, 4(Nov):933--969, 2003.

\bibitem[\protect\citeauthoryear{Ghaddar and Naoum-Sawaya}{2018}]{ghaddar2018high}
Bissan Ghaddar and Joe Naoum-Sawaya.
\newblock High dimensional data classification and feature selection using support vector machines.
\newblock {\em European Journal of Operational Research}, 265(3):993--1004, 2018.

\bibitem[\protect\citeauthoryear{Jim{\'e}nez-Cordero \bgroup \em et al.\egroup }{2021}]{jimenez2021novel}
Asunci{\'o}n Jim{\'e}nez-Cordero, Juan~Miguel Morales, and Salvador Pineda.
\newblock A novel embedded min-max approach for feature selection in nonlinear support vector machine classification.
\newblock {\em European Journal of Operational Research}, 293(1):24--35, 2021.

\bibitem[\protect\citeauthoryear{Joachims}{2002}]{joachims2002optimizing}
Thorsten Joachims.
\newblock Optimizing search engines using clickthrough data.
\newblock In {\em Proceedings of the eighth ACM SIGKDD international conference on Knowledge discovery and data mining}, pages 133--142. ACM, 2002.

\bibitem[\protect\citeauthoryear{Khoshgoftaar \bgroup \em et al.\egroup }{2007}]{khoshgoftaar2007learning}
Taghi~M Khoshgoftaar, Chris Seiffert, Jason Van~Hulse, Amri Napolitano, and Andres Folleco.
\newblock Learning with limited minority class data.
\newblock In {\em Sixth International Conference on Machine Learning and Applications (ICMLA 2007)}, pages 348--353. IEEE, 2007.

\bibitem[\protect\citeauthoryear{King}{2009}]{dlib09}
Davis~E. King.
\newblock Dlib-ml: A machine learning toolkit.
\newblock {\em Journal of Machine Learning Research}, 10:1755--1758, 2009.

\bibitem[\protect\citeauthoryear{Leevy \bgroup \em et al.\egroup }{2018}]{leevy2018survey}
Joffrey~L Leevy, Taghi~M Khoshgoftaar, Richard~A Bauder, and Naeem Seliya.
\newblock A survey on addressing high-class imbalance in big data.
\newblock {\em Journal of Big Data}, 5(1):1--30, 2018.

\bibitem[\protect\citeauthoryear{Li \bgroup \em et al.\egroup }{2017}]{li2017feature}
Jundong Li, Kewei Cheng, Suhang Wang, Fred Morstatter, Robert~P Trevino, Jiliang Tang, and Huan Liu.
\newblock Feature selection: A data perspective.
\newblock {\em ACM computing surveys (CSUR)}, 50(6):1--45, 2017.

\bibitem[\protect\citeauthoryear{Ling \bgroup \em et al.\egroup }{2003}]{ling2003auc}
Charles~X Ling, Jin Huang, Harry Zhang, et~al.
\newblock {AUC}: a statistically consistent and more discriminating measure than accuracy.
\newblock In {\em Ijcai}, volume~3, pages 519--524, 2003.

\bibitem[\protect\citeauthoryear{Ming and Ding}{2019}]{ming2019robust}
Di~Ming and Chris Ding.
\newblock Robust flexible feature selection via exclusive l21 regularization.
\newblock In {\em Proceedings of the 28th international joint conference on artificial intelligence}, pages 3158--3164, 2019.

\bibitem[\protect\citeauthoryear{Norton and Uryasev}{2019}]{norton2019maximization}
Matthew Norton and Stan Uryasev.
\newblock Maximization of {AUC} and buffered {AUC} in binary classification.
\newblock {\em Mathematical Programming}, 174(1):575--612, 2019.

\bibitem[\protect\citeauthoryear{Ozcan \bgroup \em et al.\egroup }{2024}]{ozcan2023column}
Erhan~C Ozcan, Berk G{\"o}rg{\"u}l{\"u}, and Mustafa~G Baydogan.
\newblock Column generation-based prototype learning for optimizing area under the receiver operating characteristic curve.
\newblock {\em European Journal of Operational Research}, 314(1):297--307, 2024.

\bibitem[\protect\citeauthoryear{Rakotomamonjy}{2004}]{rakotomamonjy2004optimizing}
Alain Rakotomamonjy.
\newblock Optimizing area under {ROC} curve with {SVM}s.
\newblock In {\em ROCAI}, pages 71--80, 2004.

\bibitem[\protect\citeauthoryear{Van~Hulse \bgroup \em et al.\egroup }{2007}]{van2007experimental}
Jason Van~Hulse, Taghi~M Khoshgoftaar, and Amri Napolitano.
\newblock Experimental perspectives on learning from imbalanced data.
\newblock In {\em Proceedings of the 24th international conference on Machine learning}, pages 935--942, 2007.

\bibitem[\protect\citeauthoryear{Wang \bgroup \em et al.\egroup }{2012}]{wang2012using}
Xiaoguang Wang, Hang Shao, Nathalie Japkowicz, Stan Matwin, Xuan Liu, Alex Bourque, and Bao Nguyen.
\newblock Using svm with adaptively asymmetric misclassification costs for mine-like objects detection.
\newblock In {\em 2012 11th International Conference on Machine Learning and Applications}, volume~2, pages 78--82. IEEE, 2012.

\bibitem[\protect\citeauthoryear{Yan \bgroup \em et al.\egroup }{2003}]{yan2003optimizing}
Lian Yan, Robert~H Dodier, Michael Mozer, and Richard~H Wolniewicz.
\newblock Optimizing classifier performance via an approximation to the wilcoxon-mann-whitney statistic.
\newblock In {\em Proceedings of the 20th International Conference on Machine Learning (ICML-03)}, pages 848--855, 2003.

\bibitem[\protect\citeauthoryear{Zhang and Zhou}{2013}]{zhang2013analysis}
Li~Zhang and WeiDa Zhou.
\newblock Analysis of programming properties and the row--column generation method for 1-norm support vector machines.
\newblock {\em Neural Networks}, 48:32--43, 2013.

\bibitem[\protect\citeauthoryear{Zhang \bgroup \em et al.\egroup }{2015}]{zhang2015dissimilarity}
Xueying Zhang, Qinbao Song, Guangtao Wang, Kaiyuan Zhang, Liang He, and Xiaolin Jia.
\newblock A dissimilarity-based imbalance data classification algorithm.
\newblock {\em Applied Intelligence}, 42:544--565, 2015.

\end{thebibliography}

\clearpage

\onecolumn

\section*{Appendix A: Dataset Characteristics}\label{app:dataset_characteristics}

\renewcommand{\thetable}{A.1}
\begin{table*}[!h]
\centering
\caption{Summary of the Datasets}
\label{tab:dataset}
\renewcommand{\arraystretch}{.9}
\resizebox{\textwidth}{!}{
\begin{tabular}{lcccc}
\hline
\textbf{Dataset}                          & \textbf{\begin{tabular}[c]{@{}c@{}}\# of Train \\ Instances\end{tabular}} & \textbf{\begin{tabular}[c]{@{}c@{}}\# of Test \\ Instances\end{tabular}} & \textbf{\# of Features} & \textbf{\begin{tabular}[c]{@{}c@{}}Class Imbalance Ratio \\ (\# of Class 1 / \# of Class -1)\end{tabular}} \\ \hline
\textbf{abalone-21\_vs\_8}                & 435                               & 146                           & 9                       & 0.02                                                                                                   \\
\textbf{abalone-3\_vs\_11}                & 376                               & 126                           & 9                       & 0.03                                                                                                   \\
\textbf{abalone9-18}                      & 548                               & 183                           & 9                       & 0.06                                                                                                   \\
\textbf{cleveland-0\_vs\_4}               & 129                               & 44                            & 13                      & 0.08                                                                                                    \\
\textbf{dermatology-6}                    & 268                               & 90                            & 34                      & 0.06                                                                                                   \\
\textbf{ecoli-0\_vs\_1}                   & 165                               & 55                            & 7                       & 0.54                                                                                                   \\
\textbf{ecoli-0-1\_vs\_2-3-5}             & 183                               & 61                            & 7                       & 0.11                                                                                                   \\
\textbf{ecoli-0-1-3-7\_vs\_2-6}           & 210                               & 71                            & 7                       & 0.02                                                                                                   \\
\textbf{ecoli-0-1-4-7\_vs\_2-3-5-6}       & 252                               & 84                            & 7                       & 0.09                                                                                                   \\
\textbf{ecoli-0-2-3-4\_vs\_5}             & 151                               & 51                            & 7                       & 0.11                                                                                                    \\
\textbf{ecoli-0-2-6-7\_vs\_3-5}           & 168                               & 56                            & 7                       & 0.11                                                                                                   \\
\textbf{ecoli-0-3-4\_vs\_5}               & 150                               & 50                            & 7                       & 0.11                                                                                                   \\
\textbf{ecoli-0-3-4-6\_vs\_5}             & 153                               & 52                            & 7                       & 0.11                                                                                                   \\
\textbf{ecoli-0-3-4-7\_vs\_5-6}           & 192                               & 65                            & 7                       & 0.11                                                                                                   \\
\textbf{ecoli-0-6-7\_vs\_3-5}             & 166                               & 56                            & 7                       & 0.11                                                                                                       \\
\textbf{ecoli1}                           & 252                               & 84                            & 7                       & 0.30                                                                                                   \\
\textbf{ecoli2}                           & 252                               & 84                            & 7                       & 0.18                                                                                                   \\
\textbf{ecoli4}                           & 252                               & 84                            & 7                       & 0.06                                                                                                   \\
\textbf{flare-F}                          & 799                               & 267                           & 17                      & 0.04                                                                                                   \\
\textbf{glass0}                           & 160                               & 54                            & 9                       & 0.49                                                                                                   \\
\textbf{glass-0-1-2-3\_vs\_4-5-6}         & 160                               & 54                            & 9                       & 0.31                                                                                                   \\
\textbf{glass-0-1-4-6\_vs\_2}             & 153                               & 52                            & 9                       & 0.09                                                                                                   \\
\textbf{glass-0-1-5\_vs\_2}               & 129                               & 43                            & 9                       & 0.11                                                                                                   \\
\textbf{glass-0-1-6\_vs\_2}               & 144                               & 48                            & 9                       & 0.10                                                                                                   \\
\textbf{glass-0-1-6\_vs\_5}               & 138                               & 46                            & 9                       & 0.05                                                                                                   \\
\textbf{glass-0-4\_vs\_5}                 & 69                                & 23                            & 9                       & 0.11                                                                                                   \\
\textbf{glass-0-6\_vs\_5}                 & 81                                & 27                            & 9                       & 0.09                                                                                                   \\
\textbf{glass1}                           & 160                               & 54                            & 9                       & 0.55                                                                                                   \\
\textbf{glass4}                           & 160                               & 54                            & 9                       & 0.06                                                                                                   \\
\textbf{glass5}                           & 160                               & 54                            & 9                       & 0.04                                                                                                   \\
\textbf{glass6}                           & 160                               & 54                            & 9                       & 0.16                                                                                                   \\
\textbf{haberman}                         & 229                               & 77                            & 3                       & 0.36                                                                                                       \\
\textbf{ionosphere}                       & 263                               & 88                            & 34                      & 0.56                                                                                                       \\
\textbf{iris0}                            & 112                               & 38                            & 4                       & 0.50                                                                                                        \\
\textbf{kr-vs-k-zero\_vs\_eight}          & 1095                              & 365                           & 20                      & 0.02                                                                                                   \\
\textbf{led7digit-0-2-4-5-6-7-8-9\_vs\_1} & 332                               & 111                           & 7                       & 0.09                                                                                                   \\
\textbf{lymphography-normal-fibrosis}     & 111                               & 37                            & 32                      & 0.04                                                                                                   \\
\textbf{new-thyroid1}                     & 161                               & 54                            & 5                       & 0.19                                                                                                   \\
\textbf{new-thyroid2}                     & 161                               & 54                            & 5                       & 0.19                                                                                                   \\
\textbf{page-blocks-1-3\_vs\_4}           & 354                               & 118                           & 10                      & 0.06                                                                                                   \\
\textbf{parkinsons}                       & 146                               & 49                            & 22                      & 0.33                                                                                                   \\
\textbf{pima}                             & 576                               & 192                           & 8                       & 0.54                                                                                                      \\
\hline
\end{tabular}}
\end{table*}

\renewcommand{\thetable}{A.1}
\begin{table*}[!h]
\centering
\caption{Summary of the Datasets (cont.)}
\renewcommand{\arraystretch}{.9}
\resizebox{\textwidth}{!}{
\begin{tabular}{lcccc}
\hline
\textbf{Dataset}                          & \textbf{\begin{tabular}[c]{@{}c@{}}\# of Train \\ Instances\end{tabular}} & \textbf{\begin{tabular}[c]{@{}c@{}}\# of Test \\ Instances\end{tabular}} & \textbf{\# of Features} & \textbf{\begin{tabular}[c]{@{}c@{}}Class Imbalance Ratio \\ (\# of Class 1 / \# of Class -1)\end{tabular}} \\ \hline
\textbf{poker-8\_vs\_6}                   & 1107                              & 370                           & 10                      & 0.02                                                                                                   \\
\textbf{poker-8-9\_vs\_6}                 & 1113                              & 372                           & 10                      & 0.02                                                                                                   \\
\textbf{poker-9\_vs\_7}                   & 183                               & 61                            & 10                      & 0.03                                                                                                   \\
\textbf{shuttle-6\_vs\_2-3}               & 172                               & 58                            & 9                       & 0.04                                                                                                   \\
\textbf{shuttle-c2-vs-c4}                 & 96                                & 33                            & 9                       & 0.05                                                                                                    \\
\textbf{sonar}                            & 156                               & 52                            & 60                      & 0.87                                                                                                   \\
\textbf{survival}                         & 229                               & 77                            & 3                       & 0.36                                                                                                       \\
\textbf{vehicle1}                         & 634                               & 212                           & 18                      & 0.34                                                                                                   \\
\textbf{vehicle3}                         & 634                               & 212                           & 18                      & 0.33                                                                                                   \\
\textbf{votes}                            & 326                               & 109                           & 32                      & 0.63                                                                                                   \\
\textbf{vowel0}                           & 741                               & 247                           & 13                      & 0.10                                                                                                   \\
\textbf{winequality-red-3\_vs\_5}         & 518                               & 173                           & 11                      & 0.01                                                                                                   \\
\textbf{winequality-red-4}                & 1199                              & 400                           & 11                      & 0.03                                                                                                   \\
\textbf{winequality-red-8\_vs\_6}         & 492                               & 164                           & 11                      & 0.03                                                                                                   \\
\textbf{winequality-red-8\_vs\_6-7}       & 641                               & 214                           & 11                      & 0.02                                                                                                   \\
\textbf{winequality-white-3\_vs\_7}       & 675                               & 225                           & 11                      & 0.02                                                                                                   \\
\textbf{winequality-white-9\_vs\_4}       & 126                               & 42                            & 11                      & 0.03                                                                                                   \\
\textbf{wisconsin}                        & 512                               & 171                           & 9                       & 0.54                                                                                                   \\
\textbf{yeast-0-2-5-6\_vs\_3-7-8-9}       & 753                               & 251                           & 8                       & 0.11                                                                                                   \\
\textbf{yeast-0-2-5-7-9\_vs\_3-6-8}       & 753                               & 251                           & 8                       & 0.11                                                                                                   \\
\textbf{yeast-0-3-5-9\_vs\_7-8}           & 379                               & 127                           & 8                       & 0.11                                                                                                   \\
\textbf{yeast-0-5-6-7-9\_vs\_4}           & 396                               & 132                           & 8                       & 0.11                                                                                                   \\
\textbf{yeast-1-2-8-9\_vs\_7}             & 710                               & 237                           & 8                       & 0.03                                                                                                   \\
\textbf{yeast-1-4-5-8\_vs\_7}             & 519                               & 174                           & 8                       & 0.04                                                                                                   \\
\textbf{yeast-2\_vs\_4}                   & 385                               & 129                           & 8                       & 0.11                                                                                                   \\
\textbf{yeast-2\_vs\_8}                   & 361                               & 121                           & 8                       & 0.04                                                                                                    \\
\textbf{yeast3}                           & 1113                              & 371                           & 8                       & 0.12                                                                                                   \\
\textbf{yeast4}                           & 1113                              & 371                           & 8                       & 0.03                                                                                                    \\
\textbf{yeast5}                           & 1113                              & 371                           & 8                       & 0.03                                                                                                   \\
\textbf{yeast6}                           & 1113                              & 371                           & 8                       & 0.02                                                                                                   \\
\textbf{zoo-3}                            & 75                                & 26                            & 16                      & 0.05                                                                                                   \\ \hline
\end{tabular}}
\end{table*}

\clearpage

\section*{Appendix B: The Details of the Tuned Parameters}\label{app:parameters}

\renewcommand{\thetable}{B.1}
\begin{table*}[!h]
\centering
\caption{Values considered in parameter tuning for Smooth Ranking-CG Prototype and the benchmark methods.}
\label{tab:hypertune}
\resizebox{\textwidth}{!}{
\begin{tabular}{ll}
\hline
\textbf{Method}                         & \textbf{Parameters}                                                                                                                                                                                                                                                    \\ \hline
\textbf{Smooth Ranking-CG Prototype}    & \begin{tabular}[c]{@{}l@{}}$C \in \Bigl\{10^{-6}, 5\times 10^{-6}, 10^{-5}, 5\times 10^{-5}, 10^{-4}, 5\times 10^{-4}, 10^{-3}, 5\times 10^{-3}, 10^{-2}, 5\times 10^{-2}, 10^{-1}, 5\times 10^{-1}, 10^{0}\Bigr\}$\end{tabular} \\
\textbf{Ranking-CG Prototype}           & \begin{tabular}[c]{@{}l@{}}$\alpha \in \Bigl\{10^{-5}, 5\times 10^{-5}, 10^{-4}, 2.5 \times 10^{-4}, 5 \times 10^{-4}, 7.5 \times 10^{-4}, 10^{-3}, 2.5\times 10^{-3},  5\times 10^{-3}, 7.5\times 10^{-3}, 10^{-2}, 2.5 \times 10^{-2}, 5 \times 10^{-2}\Bigr\}$\end{tabular} \\
\textbf{Ranking SVM}                    & \begin{tabular}[c]{@{}l@{}}$C \in \Bigl\{10^{-3}, 5\times 10^{-3}, 10^{-2}, 5\times 10^{-2}, 10^{-1}, 5\times 10^{-1}, 1, 5, 10^{-1}, 5\times 10^{-1}, 10^{2}, 5\times 10^{2}, 10^{3}, 5\times 10^{3}\Bigr\}$\end{tabular}                                                    \\
\textbf{$L_1$ Ranking}                  & \begin{tabular}[c]{@{}l@{}}$C \in \Bigl\{10^{-3}, 5\times 10^{-3}, 10^{-2}, 5\times 10^{-2}, 10^{-1}, 5\times 10^{-1}, 1, 5, 10^{-1}, 5\times 10^{-1}, 10^{2}, 5\times 10^{2}, 10^{3}, 5\times 10^{3}\Bigr\}$\end{tabular}                                                    \\
\textbf{$L_{\infty}$ Ranking}           & \begin{tabular}[c]{@{}l@{}}$C \in \Bigl\{10^{-3}, 5\times 10^{-3}, 10^{-2}, 5\times 10^{-2}, 10^{-1}, 5\times 10^{-1}, 1, 5, 10^{-1}, 5\times 10^{-1}, 10^{2}, 5\times 10^{2}, 10^{3}, 5\times 10^{3}\Bigr\}$\end{tabular}                                                    \\ \hline
\end{tabular}
}
\end{table*}

\clearpage

\section*{Appendix C: Detailed Results}\label{app:detailed_results}
\renewcommand{\thetable}{C.1}
\begin{table*}[!h]
\caption{Out-of-sample AUC for Smooth Ranking-CG Prototype, Ranking-CG Prototype, $L_1$ Ranking, $L_{\infty}$ Ranking, and Ranking-SVM for 74 datasets.}
\label{tab:app_detailed}
\renewcommand{\arraystretch}{1}
\scalebox{0.83}{
\begin{tabular}{lccccc}
\hline
\textbf{Dataset}                    & \textbf{Smooth Ranking-CG Prototype} & \textbf{Ranking-CG Prototype} & \textbf{$L_1$ Ranking} & \textbf{$L_{\infty}$ Ranking} & \textbf{Ranking SVM} \\ \hline 
\textbf{abalone-21\_vs\_8}                &           0.975 &                 1.000 &    0.949 &       0.968 &    0.991 \\
\textbf{abalone-3\_vs\_11}                &           1.000 &                 1.000 &    1.000 &       1.000 &    1.000 \\
\textbf{abalone9-18}                    &           0.960 &                 0.939 &    0.940 &       0.879 &    0.929 \\
\textbf{cleveland-0\_vs\_4}               &           0.935 &                 0.984 &    0.967 &       0.967 &    0.959 \\
\textbf{dermatology-6}                  &           1.000 &                 1.000 &    1.000 &       1.000 &    1.000 \\
\textbf{ecoli-0-1-3-7\_vs\_2-6}           &           1.000 &                 1.000 &    1.000 &       1.000 &    1.000 \\
\textbf{ecoli-0-1-4-7\_vs\_2-3-5-6}       &           0.976 &                 0.976 &    0.952 &       0.881 &    0.939 \\
\textbf{ecoli-0-1\_vs\_2-3-5}             &           0.967 &                 0.942 &    0.958 &       0.955 &    0.967 \\
\textbf{ecoli-0-2-3-4\_vs\_5}             &           0.991 &                 1.000 &    1.000 &       1.000 &    0.991 \\
\textbf{ecoli-0-2-6-7\_vs\_3-5}           &           0.893 &                 0.837 &    0.867 &       0.910 &    0.890 \\
\textbf{ecoli-0-3-4-6\_vs\_5}             &           0.774 &                 0.826 &    0.800 &       0.898 &    0.906 \\
\textbf{ecoli-0-3-4-7\_vs\_5-6}           &           1.000 &                 1.000 &    1.000 &       1.000 &    1.000 \\
\textbf{ecoli-0-3-4\_vs\_5}               &           0.991 &                 1.000 &    0.991 &       1.000 &    0.987 \\
\textbf{ecoli-0-6-7\_vs\_3-5}             &           0.937 &                 0.893 &    0.887 &       0.930 &    0.910 \\
\textbf{ecoli-0\_vs\_1  }                 &           0.991 &                 0.993 &    0.994 &       0.994 &    0.991 \\
\textbf{ecoli1 }                        &           0.943 &                 0.950 &    0.940 &       0.959 &    0.947 \\
\textbf{ecoli2  }                       &           0.948 &                 0.945 &    0.938 &       0.938 &    0.940 \\
\textbf{ecoli4  }                       &           1.000 &                 1.000 &    1.000 &       1.000 &    1.000 \\
\textbf{flare-F    }                    &           0.905 &                 0.904 &    0.863 &       0.838 &    0.874 \\
\textbf{glass-0-1-2-3\_vs\_4-5-6}         &           0.955 &                 0.977 &    0.953 &       0.931 &    0.968 \\
\textbf{glass-0-1-4-6\_vs\_2 }            &           0.812 &                 0.818 &    0.771 &       0.771 &    0.755 \\
\textbf{glass-0-1-5\_vs\_2 }              &           0.897 &                 0.885 &    0.756 &       0.885 &    0.878 \\
\textbf{glass-0-1-6\_vs\_2 }              &           0.653 &                 0.710 &    0.716 &       0.670 &    0.699 \\
\textbf{glass-0-1-6\_vs\_5 }              &           0.500 &                 0.545 &    0.466 &       0.920 &    0.989 \\
\textbf{glass-0-4\_vs\_5  }               &           0.643 &                 0.976 &    0.714 &       1.000 &    1.000 \\
\textbf{glass-0-6\_vs\_5 }                &           0.980 &                 1.000 &    1.000 &       1.000 &    1.000 \\
\textbf{glass0}                         &           0.887 &                 0.861 &    0.880 &       0.840 &    0.877 \\
\textbf{glass1 }                        &           0.841 &                 0.847 &    0.853 &       0.857 &    0.880 \\
\textbf{glass4 }                        &           0.954 &                 0.980 &    0.902 &       0.961 &    0.928 \\
\textbf{glass5 }                        &           1.000 &                 1.000 &    1.000 &       1.000 &    1.000 \\
\textbf{glass6}                         &           0.945 &                 0.936 &    0.888 &       0.936 &    0.866 \\
\hline
\end{tabular}}
\end{table*}

\renewcommand{\thetable}{C.1}
\begin{table*}[!h]
\caption{Out-of-sample AUC for Smooth Ranking-CG Prototype, Ranking-CG Prototype, $L_1$ Ranking, $L_{\infty}$ Ranking, and Ranking-SVM for 74 datasets. (cont.)}
\renewcommand{\arraystretch}{1}
\scalebox{0.83}{
\begin{tabular}{lccccc}
\hline
\textbf{Dataset}                    & \textbf{Smooth Ranking-CG Prototype} & \textbf{Ranking-CG Prototype} & \textbf{$L_1$ Ranking} & \textbf{$L_{\infty}$ Ranking} & \textbf{Ranking SVM} \\ \hline 
\textbf{haberman                   }    &           0.634 &                 0.638 &    0.619 &       0.587 &    0.625 \\
\textbf{ionosphere                }     &           0.980 &                 0.980 &    0.994 &       0.993 &    0.990 \\
\textbf{iris0                    }      &           1.000 &                 1.000 &    1.000 &       1.000 &    1.000 \\
\textbf{kr-vs-k-zero\_vs\_eight    }      &           0.994 &                 0.998 &    1.000 &       1.000 &    1.000 \\
\textbf{led7digit-0-2-4-5-6-7-8-9\_vs\_1} &           0.973 &                 0.977 &    0.983 &       0.974 &    0.973 \\
\textbf{lymphography-normal-fibrosis}   &           0.986 &                 0.986 &    0.986 &       0.971 &    0.929 \\
\textbf{new-thyroid1             }      &           0.669 &                 0.704 &    0.995 &       0.817 &    0.906 \\
\textbf{new-thyroid2            }       &           0.990 &                 0.995 &    0.998 &       0.995 &    0.998 \\
\textbf{page-blocks-1-3\_vs\_4 }          &           0.986 &                 0.569 &    0.568 &       0.569 &    0.777 \\
\textbf{parkinsons            }         &           0.842 &                 0.854 &    0.865 &       0.854 &    0.858 \\
\textbf{pima                 }          &           0.830 &                 0.847 &    0.839 &       0.759 &    0.841 \\
\textbf{poker-8-9\_vs\_6           }      &           0.523 &                 0.536 &    0.597 &       0.496 &    0.555 \\
\textbf{poker-8\_vs\_6            }       &           0.645 &                 0.704 &    0.676 &       0.463 &    0.527 \\
\textbf{poker-9\_vs\_7           }        &           0.873 &                 0.983 &    0.805 &       0.983 &    1.000 \\
\textbf{shuttle-6\_vs\_2-3      }         &           1.000 &                 1.000 &    1.000 &       1.000 &    1.000 \\
\textbf{shuttle-c2-vs-c4       }        &           1.000 &                 1.000 &    0.968 &       0.935 &    0.935 \\
\textbf{sonar                    }      &           0.899 &                 0.902 &    0.930 &       0.966 &    0.958 \\
\textbf{survival                }       &           0.659 &                 0.697 &    0.683 &       0.657 &    0.690 \\
\textbf{vehicle1               }        &           0.660 &                 0.904 &    0.916 &       0.908 &    0.913 \\
\textbf{vehicle3              }         &           0.748 &                 0.802 &    0.853 &       0.864 &    0.852 \\
\textbf{votes                }          &           0.997 &                 0.991 &    0.984 &       0.986 &    0.993 \\
\textbf{vowel0              }           &           0.998 &                 1.000 &    1.000 &       1.000 &    1.000 \\
\textbf{winequality-red-3\_vs\_5  }       &           0.763 &                 0.843 &    0.776 &       0.778 &    0.794 \\
\textbf{winequality-red-4        }      &           0.561 &                 0.524 &    0.461 &       0.410 &    0.452 \\
\textbf{winequality-red-8\_vs\_6     }    &           0.649 &                 0.669 &    0.668 &       0.713 &    0.649 \\
\textbf{winequality-red-8\_vs\_6-7   }    &           0.412 &                 0.422 &    0.424 &       0.389 &    0.363 \\
\textbf{winequality-white-3\_vs\_7   }    &           0.884 &                 0.943 &    0.966 &       0.866 &    0.940 \\
\textbf{winequality-white-9\_vs\_4   }    &           0.683 &                 0.537 &    0.561 &       0.537 &    0.537 \\
\textbf{wisconsin                   }   &           0.998 &                 0.997 &    0.998 &       0.998 &    0.998 \\
\textbf{yeast-0-2-5-6\_vs\_3-7-8-9 }      &           0.851 &                 0.830 &    0.817 &       0.740 &    0.822 \\
\textbf{yeast-0-2-5-7-9\_vs\_3-6-8}       &           0.925 &                 0.929 &    0.950 &       0.928 &    0.941 \\
\textbf{yeast-0-3-5-9\_vs\_7-8  }         &           0.719 &                 0.734 &    0.725 &       0.707 &    0.746 \\
\textbf{yeast-0-5-6-7-9\_vs\_4 }          &           0.849 &                 0.879 &    0.867 &       0.899 &    0.871 \\
\textbf{yeast-1-2-8-9\_vs\_7 }            &           0.820 &                 0.941 &    0.920 &       0.810 &    0.905 \\
\textbf{yeast-1-4-5-8\_vs\_7}             &           0.727 &                 0.696 &    0.645 &       0.786 &    0.601 \\
\textbf{yeast-2\_vs\_4 }                  &           0.979 &                 0.975 &    0.971 &       0.954 &    0.982 \\
\textbf{yeast-2\_vs\_8}                   &           0.721 &                 0.872 &    0.910 &       0.845 &    0.760 \\
\textbf{yeast3       }                  &           0.959 &                 0.955 &    0.965 &       0.936 &    0.972 \\
\textbf{yeast4      }                   &           0.910 &                 0.927 &    0.930 &       0.912 &    0.914 \\
\textbf{yeast5     }                    &           0.986 &                 0.985 &    0.992 &       0.987 &    0.991 \\
\textbf{yeast6    }                     &           0.948 &                 0.944 &    0.930 &       0.911 &    0.957 \\
\textbf{zoo-3    }                      &           0.720 &                 0.960 &    1.000 &       1.000 &    1.000 \\
\hline
\end{tabular}}
\end{table*}

\renewcommand{\thetable}{C.2}
\begin{table*}[!h]
\caption{Number of features with coefficients' magnitude $\geq 0.001$ for Smooth Ranking-CG Prototype, Ranking-CG Prototype, $L_1$ Ranking, $L_{\infty}$ Ranking, and Ranking-SVM for 74 datasets.}
\renewcommand{\arraystretch}{1}
\scalebox{0.83}{
\begin{tabular}{lccccc}
\hline
\textbf{Dataset}                    & \textbf{Smooth Ranking-CG Prototype} & \textbf{Ranking-CG Prototype} & \textbf{$L_1$ Ranking} & \textbf{$L_{\infty}$ Ranking} & \textbf{Ranking SVM} \\ \hline 
\textbf{abalone-21\_vs\_8          }      &               6 &                    13 &       12 &         435 &      433 \\
\textbf{abalone-3\_vs\_11          }      &               2 &                     3 &        4 &         376 &      351 \\
\textbf{abalone9-18                }    &               6 &                     7 &       79 &         548 &      546 \\
\textbf{cleveland-0\_vs\_4         }      &               3 &                    29 &        6 &         129 &       32 \\
\textbf{dermatology-6              }    &               3 &                     7 &       10 &         268 &      152 \\
\textbf{ecoli-0-1-3-7\_vs\_2-6     }      &               5 &                     6 &        6 &         209 &       21 \\
\textbf{ecoli-0-1-4-7\_vs\_2-3-5-6}       &               6 &                     6 &       18 &         252 &       73 \\
\textbf{ecoli-0-1\_vs\_2-3-5     }        &               5 &                    13 &        7 &         183 &       18 \\
\textbf{ecoli-0-2-3-4\_vs\_5     }        &               4 &                     9 &       23 &         150 &       18 \\
\textbf{ecoli-0-2-6-7\_vs\_3-5   }        &               4 &                    10 &        2 &         168 &       26 \\
\textbf{ecoli-0-3-4-6\_vs\_5     }        &               5 &                     8 &       13 &         153 &       17 \\
\textbf{ecoli-0-3-4-7\_vs\_5-6   }        &               5 &                    11 &        7 &         192 &      151 \\
\textbf{ecoli-0-3-4\_vs\_5       }        &               4 &                     6 &        6 &         150 &       16 \\
\textbf{ecoli-0-6-7\_vs\_3-5     }        &               4 &                     7 &        4 &         166 &       25 \\
\textbf{ecoli-0\_vs\_1           }        &               2 &                     7 &       15 &         165 &      165 \\
\textbf{ecoli1                   }      &               4 &                     9 &       13 &         252 &      250 \\
\textbf{ecoli2                   }      &               4 &                    15 &       25 &         252 &      252 \\
\textbf{ecoli4                   }      &               3 &                     7 &        8 &         252 &      248 \\
\textbf{flare-F                  }      &               5 &                     6 &        9 &         799 &      757 \\
\textbf{glass-0-1-2-3\_vs\_4-5-6}         &               3 &                    13 &       31 &         159 &      157 \\
\textbf{glass-0-1-4-6\_vs\_2  }           &               4 &                     8 &       32 &         153 &      153 \\
\textbf{glass-0-1-5\_vs\_2   }            &              10 &                    29 &        7 &         129 &      129 \\
\textbf{glass-0-1-6\_vs\_2  }             &               6 &                     8 &       30 &         144 &      143 \\
\textbf{glass-0-1-6\_vs\_5 }              &               3 &                     4 &        2 &         138 &      135 \\
\textbf{glass-0-4\_vs\_5 }                &               2 &                     2 &        6 &          69 &       68 \\
\textbf{glass-0-6\_vs\_5}                 &               3 &                     3 &       13 &          81 &       80 \\
\textbf{glass0    }                     &               8 &                    33 &       34 &         160 &      158 \\
\textbf{glass1   }                      &               7 &                    68 &       52 &         160 &      158 \\
\textbf{glass4  }                       &               4 &                     7 &        5 &         159 &      144 \\
\textbf{glass5 }                        &               5 &                     5 &        8 &         160 &      148 \\
\textbf{glass6}                         &               3 &                     7 &       14 &         160 &      159 \\
\hline
\end{tabular}}
\end{table*}

\renewcommand{\thetable}{C.2}
\begin{table*}[!h]
\caption{Number of features with coefficients' magnitude $\geq 0.001$ for Smooth Ranking-CG Prototype, Ranking-CG Prototype, $L_1$ Ranking, $L_{\infty}$ Ranking, and Ranking-SVM for 74 datasets. (cont.)}
\renewcommand{\arraystretch}{1}
\scalebox{0.83}{
\begin{tabular}{lccccc}
\hline
\textbf{Dataset}                    & \textbf{Smooth Ranking-CG Prototype} & \textbf{Ranking-CG Prototype} & \textbf{$L_1$ Ranking} & \textbf{$L_{\infty}$ Ranking} & \textbf{Ranking SVM} \\ \hline 
\textbf{haberman                     }  &               5 &                    14 &       34 &         229 &      220 \\
\textbf{ionosphere                  }   &               5 &                    11 &       58 &         262 &      262 \\
\textbf{iris0                      }    &               2 &                     2 &        5 &         112 &      107 \\
\textbf{kr-vs-k-zero\_vs\_eight   }       &               6 &                    11 &       35 &        1095 &     1069 \\
\textbf{led7digit-0-2-4-5-6-7-8-9\_vs\_1} &               3 &                    15 &        6 &         332 &      316 \\
\textbf{lymphography-normal-fibrosis}   &               3 &                     3 &        7 &         111 &      104 \\
\textbf{new-thyroid1              }     &               4 &                     7 &       10 &         161 &      129 \\
\textbf{new-thyroid2             }      &               4 &                     4 &        9 &         161 &      148 \\
\textbf{page-blocks-1-3\_vs\_4  }         &               4 &                    67 &       11 &         354 &        0 \\
\textbf{parkinsons                  }   &               9 &                    59 &       61 &         146 &      132 \\
\textbf{pima                       }    &               7 &                    34 &       52 &         576 &      192 \\
\textbf{poker-8-9\_vs\_6          }       &              19 &                    22 &      151 &        1098 &     1029 \\
\textbf{poker-8\_vs\_6           }        &              11 &                    32 &       90 &        1090 &      385 \\
\textbf{poker-9\_vs\_7          }         &               3 &                     8 &        5 &         182 &      172 \\
\textbf{shuttle-6\_vs\_2-3     }          &               2 &                     2 &        4 &         172 &        0 \\
\textbf{shuttle-c2-vs-c4      }         &               2 &                     2 &        3 &          96 &        6 \\
\textbf{sonar                }          &               5 &                     6 &       87 &         154 &      156 \\
\textbf{survival            }           &               7 &                     6 &       19 &         229 &      214 \\
\textbf{vehicle1           }            &               3 &                   219 &      293 &         634 &      583 \\
\textbf{vehicle3          }             &               7 &                    65 &      100 &         634 &      530 \\
\textbf{votes            }              &               3 &                     6 &       31 &         326 &      308 \\
\textbf{vowel0          }               &               6 &                    14 &       32 &         741 &      681 \\
\textbf{winequality-red-3\_vs\_5   }      &               8 &                     9 &        2 &         518 &        4 \\
\textbf{winequality-red-4          }    &               3 &                   383 &      129 &        1199 &     1154 \\
\textbf{winequality-red-8\_vs\_6   }      &              13 &                    18 &        2 &         492 &       10 \\
\textbf{winequality-red-8\_vs\_6-7}       &              11 &                    70 &       19 &         641 &      190 \\
\textbf{winequality-white-3\_vs\_7    }   &               6 &                     6 &        5 &         675 &      103 \\
\textbf{winequality-white-9\_vs\_4   }    &               3 &                     9 &        3 &         126 &       50 \\
\textbf{wisconsin                   }   &               3 &                    15 &       11 &         512 &        0 \\
\textbf{yeast-0-2-5-6\_vs\_3-7-8-9 }      &               5 &                     6 &       20 &         753 &      745 \\
\textbf{yeast-0-2-5-7-9\_vs\_3-6-8}       &               6 &                    32 &       87 &         753 &      750 \\
\textbf{yeast-0-3-5-9\_vs\_7-8 }          &               4 &                     8 &      108 &         379 &      375 \\
\textbf{yeast-0-5-6-7-9\_vs\_4}           &               4 &                    40 &       26 &         396 &      395 \\
\textbf{yeast-1-2-8-9\_vs\_7 }            &              11 &                     8 &       14 &         710 &      705 \\
\textbf{yeast-1-4-5-8\_vs\_7}             &               8 &                     6 &        7 &         519 &      490 \\
\textbf{yeast-2\_vs\_4 }                  &               4 &                     6 &       28 &         385 &      383 \\
\textbf{yeast-2\_vs\_8}                   &               4 &                    21 &       40 &         361 &      359 \\
\textbf{yeast3  }                       &               4 &                    10 &       14 &        1112 &     1106 \\
\textbf{yeast4 }                        &               5 &                     6 &       12 &        1113 &     1081 \\
\textbf{yeast5}                         &               3 &                    13 &       36 &        1113 &     1092 \\
\textbf{yeast6}                         &               4 &                     6 &        6 &        1111 &     1087 \\
\textbf{zoo-3   }                       &               3 &                     3 &       11 &          75 &       70 \\
\hline
\end{tabular}}
\end{table*}

\end{document}